\documentclass{article}
\usepackage{ijcai21}

\usepackage{times}

\usepackage{soul}
\usepackage{url}
\usepackage[hidelinks]{hyperref}
\usepackage[utf8]{inputenc}
\usepackage[small]{caption}
\usepackage{graphicx}
\usepackage{amsmath}
\usepackage{booktabs}
\usepackage{amsfonts}
\usepackage{caption, subcaption} 
\usepackage{xcolor}
\usepackage[linesnumbered,ruled,vlined]{algorithm2e}

\newtheorem{theorem}{Theorem}
\newtheorem{definition}{Definition}

\SetCommentSty{mycommfont}

\SetKwInput{KwInput}{Input}                % Set the Input
\SetKwInput{KwOutput}{Output}              % set the Output

\title{Learn to Intervene: An Adaptive Learning Policy for Restless Bandits in Application to Preventive Healthcare}

\author{
Arpita Biswas$^1$\footnote{The research was conducted when A. Biswas worked at Google.}\and 
Gaurav Aggarwal$^2$\and
Pradeep Varakantham$^2$\And
Milind Tambe$^2$\\
\affiliations
$^1 $Harvard University\\
$^2$ Google
\emails
arpitabiswas@seas.harvard.edu, \{gauravaggarwal, pvarakantham, milindtambe\}@google.com
}

\begin{document}

\maketitle

\begin{abstract}
In many public health settings, it is important for patients to adhere to health programs, such as taking medications and periodic health checks. Unfortunately, beneficiaries may gradually disengage from such programs, which is detrimental to their health. 
A concrete example of gradual disengagement has been observed by an organization that carries out a free automated call-based program for spreading preventive care information among pregnant women. 
%This program is targeted towards low-income households in India.
Many women stop picking up calls after being enrolled for a few months. To avoid such disengagements, it is important to provide timely interventions.  
Such interventions are often expensive and can be provided to only a small fraction of the beneficiaries.  
We model this scenario as a restless multi-armed bandit (RMAB) problem, where each beneficiary is assumed to transition from one state to another depending on the intervention.
%provided to them. 
Moreover, since the transition probabilities are unknown {\em a priori}, we propose a Whittle index based Q-Learning mechanism and show that it converges to the optimal solution. Our method improves over existing learning-based methods for RMABs on multiple benchmarks from literature and also on the maternal healthcare dataset. 
\end{abstract}

\section{Introduction}

Preventive timely intervention (e.g., to check adherence or to provide medicines) can be used to significantly alleviate many public health issues such as diabetes~\cite{newman2018}, hypertension~\cite{brownstein2007}, tuberculosis~\cite{chang2013,rahedi2014}, HIV~\cite{christopher2011,kenya2013}, depression~\cite{lowe2004,mundorf2018}, etc. In this paper, we focus on the area of maternal health (well-being of women during pregnancy, childbirth and the post-natal period) where preventive interventions can impact millions of women. A key challenge in such preventive intervention problems is the limited amount of resources for intervention (e.g., availability of health care workers). Furthermore, the human behavior (w.r.t. taking medicines or adhering to a protocol) changes over time and to interventions, thereby requiring a strategic assignment of the limited resources to the beneficiaries most in need. 

We are specifically motivated towards improving maternal health among low-income communities in developing countries, where maternal deaths remain unacceptably high due to not having timely preventive care information~\cite{thaddeus1994too}. We work with a non-profit organization, ARMMAN~\citeyear{armman}, that provides a free call-based program for around $2.3$ million pregnant women. This is similar to other programs, such as Mom-connect (\url{https://tinyurl.com/mom-connect-jnj}). Each enrolled woman receives $140$ automated calls to equip them with critical life-saving healthcare information across $80$ weeks (pregnancy and up to one year after childbirth). Unfortunately, the engagement behavior (overall time spent listening to automated calls) changes and, for most women, the overall engagement decreases. This can have serious implications on their health. We ask the question: \textit{how do we systematically choose whom to provide interventions (personal visit by a healthcare worker) in order to maximize the overall engagement of the beneficiaries?}

Preventive intervention problems of interest in this paper are challenging, owing to multiple key reasons: i)~number of interventions are budgeted and much smaller than the total number of beneficiaries; ii)~beneficiary's engagement is uncertain and may change after a few months; iii)~post-intervention improvement of beneficiary's engagement is uncertain; iv)~intervention decisions at a time step have an impact on the state of beneficiaries and decisions to be taken at the next step.    
A relevant model for this setting is restless multi-armed bandits (RMABs). RMABs with prior knowledge of uncertainty model have been studied for health interventions~\cite{lee2019optimal,Mate2020,mate2021risk,killian2021beyond,bhattacharya2018restless}, sensor monitoring tasks~\cite{Ianello2012,Glazebrook2006}, anti-poaching patrols~\cite{Qian2016}, and uplift modeling in eCommerce platforms~\cite{gubela2019conversion}. Due to the unpredictability of human beneficiaries, it is unrealistic to know the uncertainty model {\em a priori}. Our key contribution is in designing an intervention scheme under limited budget, without relying on {\em a priori} knowledge of the uncertainty model.
%To this end, we focus on learning intervention decisions without knowledge of underlying uncertainty.

\paragraph{Contributions.} First, we represent the preventive intervention problem as a Restless Multi-Arm Bandit (RMAB), where the uncertainty model associated with beneficiary behaviors with and without intervention are not known {\em a priori}, thereby precluding the direct application of Whittle Index based methods~\cite{whittle1988restless}. Second, we develop a model-free learning method based on Q-learning, referred to as Whittle Index based Q-Learning (WIQL) that executes actions based on the difference between Q-values of active (intervention) and passive actions. We show that WIQL converges to the optimal solution asymptotically. Finally, to show that WIQL is a general approach and applicable to multiple domains, we evaluate WIQL on various examples. We then simulate the intervention problem in the context of maternal healthcare. We demonstrate that the model assumptions we make capture the problem of concern and the intervention scheme employed by WIQL significantly improves the engagement among beneficiaries compared to the existing benchmarks.

% \noindent\textbf{Contributions}: 
% Motivated by a concrete real-world domain, we represent the intervention problem as an RMAB with unknown uncertainty model, and show empirically that it addresses the problem of concern. Given the uncertainty associated with the response from beneficiaries with and without interventions, we represent the engagement dynamics as a Markov Decision Process (MDP) with three behavioral states. If the MDPs were known, one could compute the Whittle Indices~\cite{whittle1988restless} and intervene beneficiaries who have higher index values. However, in practice, human behavior may not be known a priori. Thus, we provide a Whittle index based Q-Learning (WIQL) to dynamically learn the intervention policies. Each iteration of WIQL employs current estimate of Whittle indices, which is then used to decide whom to intervene. We prove that WIQL converges to the optimal solution asymptotically under mild assumptions. To show that WIQL is a general approach and applicable to multiple domains, we evaluate WIQL on various examples. We then simulate the intervention problem in the context of maternal healthcare. We demonstrate that the intervention scheme employed by WIQL significantly improves the engagement among beneficiaries compared to the benchmark algorithms. 

\section{Related Work}
The \textit{restless multi-armed bandit} (RMAB) problem was introduced by Whittle~\citeyear{whittle1988restless}. %The paper studied the RMAB problem with the goal of maximizing the average reward in a dynamic programming framework. 
The main result involves formulating a relaxation of the problem and solving it optimally using a heuristic called \textit{Whittle Index policy}. This policy is optimal when the underlying Markov Decision Processes satisfy \textit{indexability}, which is computationally intensive to verify. Moreover, Papadimitriou and Tsitsiklis~\citeyear{papadimitriou1994complexity} established that solving RMAB is PSPACE-hard.

There are three main threads of relevant research. The first category focuses on specific classes of RMABs. Akbarzade {\em et al.}~\citeyear{Akbarzadeh2019} provide a class of bandits with ``controlled restarts'' and state-independent policies, which possess the indexability property and a Whittle index policy. Mate {\em et al.}~\citeyear{Mate2020} consider two-state bandits to model a health intervention problem, where  taking an intervention collapses the uncertainty about their current state.  Bhattacharya~\citeyear{bhattacharya2018restless} models the problem of maximizing the coverage and spread of health information as an RMAB problem and proposes a hierarchical policy. Lee {\em et al.}~\citeyear{lee2019optimal} study the problem of screening patients to maximize early-stage cancer detection under limited resource, by formulating it as a subclass of RMAB. Similarly, Glazebrook {\em et al.}~\citeyear{Glazebrook2006}, Hsu~\citeyear{Hsu2018}, Sombabu {\em et al.}~\citeyear{Sombabu2020}, Liu and Zhao~\citeyear{Liu2010} give Whittle indexability results for different subclasses of (hidden) Markov bandits. This category of research assumes that the transition and observation models are known beforehand. Instead, our focus is on providing learning methods when the transition model is unknown {\em a priori}. 

The second category contains different learning methods for RMABs. Fu {\em et al.}~\citeyear{fu2019towards} provide a Q-learning method where the Q value is defined based on the Whittle indices, states, and actions. However, they do not provide proof of convergence to optimal solution and experimentally, do not learn (near-)optimal policies. Along similar lines, Avrachenkov and Borkar \citeyear{avrachenkov2020whittle} provide a fundamental change to the Q-value definition for computing optimal whittle index policy. However, their convergence proof requires all homogeneous arms with same underlying MDPs. We provide a learning method that is not only shown to theoretically converge but also empirically outperforms the above mentioned methods on benchmark instances and also on a real (heterogeneous arms) problem setting.

The third relevant line of work is to predict adherence to a health program and effects of interventions, by formulating these as supervised learning problems. Killian {\em et al.}~\citeyear{Killian2019} use the data from \textit{99DOTS}~\cite{cross201999dots} and train a deep learning model to target patients at high risk of not adhering to the health program. On similar lines, Nisthala {\em et al.}~\citeyear{nishtala2020missed} use the engagement data of a health program and train deep learning model to predict patients who are at high risk of dropping out of the program. Son {\em et al.}~\citeyear{son2010application} use Support Vector Machine to predict adherence to medication among heart failure patients. There are many other papers~\cite{howes2012predicting,lauffenburger2018predicting} that train models on historical data for predicting adherence and provide interventions to patients who have low adherence probability. These works assume the training data to be available beforehand. In contrast, we consider the online nature of the problem where feedback is received after an intervention is provided, which in turn is used in making future decisions. %Moreover, there are scenarios when patients' adherence may not improve even after providing intervention and hence, it is important to study the effect of providing an intervention. 

\section{Preliminaries}
RMAB models various stochastic scheduling problems, where an instance is a 3-tuple ($N$, $M$, $\{MDP_i\}_{i\in N}$) with $N$ denoting the set of arms, $M$ is the budget restriction denoting how many arms can be pulled at a given time, and MDP$_i$ is the Markov Decision Process for each arm $i$. An MDP consists of a set of states $\mathcal{S}$, a set of actions $\mathcal{A}$, transition probabilities $\mathcal{P}: \mathcal{S}\times\mathcal{A} \times\mathcal{S}\mapsto [0,1]$, and reward function $R: \mathcal{S}\times\mathcal{A}\mapsto \mathbb{R}$. The action set $\mathcal{A}$ of each MDP consists of two actions: an \textit{active} action ($1$) and a \textit{passive} action ($0$). At each time step $t$, an action $A_i(t)$ is taken on an arm $i$, such that $\sum_i A_i(t) = M$. Then, each arm $i$ transitions to a new state and observes a reward, according to the underlying MDP$_i$. Let $X_i(t) \in \mathcal{S}$ and $R_i^{X_i(t)}(A_i(t))$ denote the current state and reward obtained at time $t$ respectively. Now, policy per step $\pi:X_1(t)\times\ldots\times X_N(t)\mapsto \{A_i(t)\}_{i\in N}$ can be defined as a mapping from the current states of all beneficiaries to the actions to be taken on each arm. Thus, given a policy $\pi$, the action on an arm $i$ is denoted as:
\begin{equation*}
    A_i^{\pi}(t) =
  \begin{cases}
    1   & \quad \text{if } i \text{ is selected by policy }\pi \text{ at time } t\\
    0  & \quad \text{if } i \text{ is not selected by policy }\pi \text{ at time } t
  \end{cases}
\end{equation*}
%Given a policy $\pi$, the total expected average reward is:
%\begin{equation}
%    V^{\pi} := \underset{t\rightarrow \infty}{\mathrm{lim\ inf}}\quad \frac{1}{t}\ 
%\mathbb{E}\left[\displaystyle \sum_{i\in N} \sum_{h=0}^{t-1} R_i^{X_i(h)}(A^{\pi}_i(h))\right] 
%\sum_{h=0}^{t-1} R_i(X_i(h))\right] 
%\end{equation}
The goal is to find a policy $\pi^*$ that maximizes the expected  reward until $T$ time steps, subject to the budget.
\begin{equation}
\begin{aligned}
& \underset{\pi}{\max}
& &  \frac{1}{T}\ 
\mathbb{E}\left[\displaystyle \sum_{i\in N} \sum_{t=1}^{T}  R_i^{X_i(t)}(A^{\pi}_i(t))\right]\\
& \text{s.t.} & &  \displaystyle \sum_{i\in N} A^{\pi}_i(t) = M \qquad \mbox{ for all } t=\{1,\ldots,T\} \label{eq:RMAB_LP}
\end{aligned}
\end{equation}
To deal with the computational hardness of solving this problem, an index-based heuristic policy based on the Lagrangian relaxation of the RMAB problem was proposed by Whittle~\citeyear{whittle1988restless}---at each time step $t$, an index is computed for each arm depending on the current state of the arm, transition probabilities, and reward function of its MDP. Then, the top $M$ arms with highest index values are selected.\\

\paragraph{Whittle Index-based policy.}
Whittle's relaxation is to replace the budget constraint of $M$ on the number of arms to a time-averaged constraint, i.e.,
\begin{equation*}
\frac{1}{T}\ 
\mathbb{E}\left[\displaystyle \sum_{t=1}^{T}\sum_{i \in N} A^{\pi}_i(t)\right] =  M
\end{equation*}
Further, using Lagrange's relaxation (with $\lambda$ as the Lagrange's multipliers) and dropping other constants, the objective function can be rewritten as:
\begin{equation}
  \underset{\pi}{\max}\ 
   \frac{1}{T}
\mathbb{E}\left[\displaystyle \sum_{t=1}^{T}\sum_{i \in N}\!\left(\!R_i^{X_i(t)}(A^{\pi}_i(t)) + \lambda\cdot (1\!\!-\!\!A^{\pi}_i(t))\right)\right]\label{eq:lag}
\end{equation}

Whittle showed that this problem can be decoupled and solved for each arm by computing the index $\lambda_i(Z)$ which acts like a subsidy that needs to be given to an arm $i$ at state $Z$, so that taking the action $1$ is as beneficial as taking the action $0$. %The value of $\lambda_i(Z)$ is computed as follows:
%\begin{equation}
%   \lambda_i(Z) = R_i^Z(1) - R_i^Z(0) + \displaystyle \sum _{Z'\in S} %\left(\mathcal{P}_i(Z, 1, Z')- \mathcal{P}_i(Z, 0, Z')\right)V_i(Z')  
%\end{equation}
Assuming \textit{indexability}, choosing arms with higher subsidy $\lambda_i(Z)$ leads to the optimal solution for Equation~\ref{eq:lag}. 
\begin{definition}[Indexability]\label{def:indexability}
Let $\Phi(\lambda)$ be the set of states for which it is optimal to take action $0$ when taking an active action costs $\lambda$. An arm is \textit{indexable} if $\Phi(\lambda)$ monotonically increases from $\emptyset$ to $\mathcal{S}$ when $\lambda$ increases from $-\infty$ to $+\infty$. An RMAB problem is \textit{indexable} if all the arms are indexable.
\end{definition}

% Note that it is important to have the knowledge of transition probabilities $\mathcal{P}_i(Z, a, Z')$ to compute Whittle Index and then choose the top $M$ arms according to those values. However, the transition probabilities are often unknown in most practical scenarios. 
%Since transition probabilities are often unknown in most practical scenarios, in this work, we consider the problem of learning the Whittle index while simultaneously selecting a set of best arms depending on the estimated Whittle Index. 

\section{The Model}\label{sec:model}
\begin{figure}[!th]
\centering
  \includegraphics[width=0.49\linewidth]{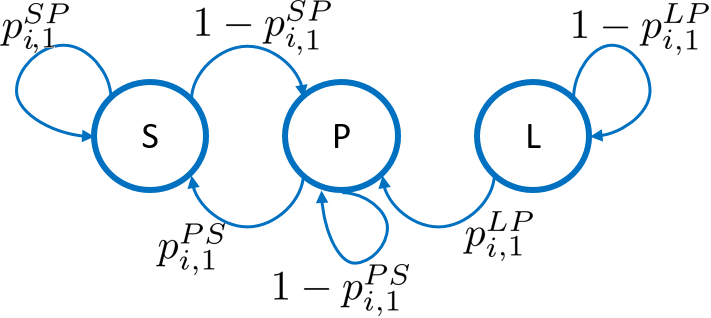}
  \includegraphics[width=0.49\linewidth]{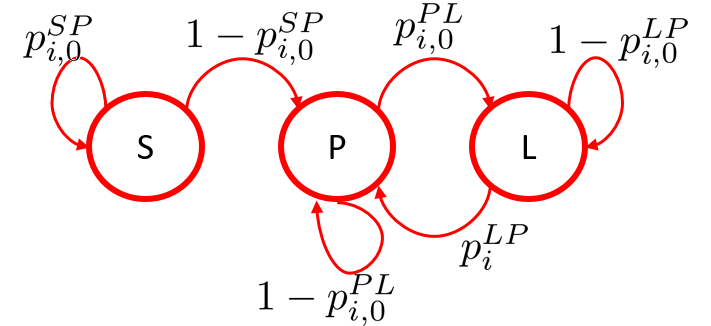}
  \caption{MDP for modeling beneficiary $i$'s engagement. Blue and red figures represent transitions $p_{i,a}^{YZ}$ from a state $Y$ to $Z$, with intervention ($a\!=\!1$) and without ($a\!=\!0$),  respectively.}
  \label{fig:MDP}
\end{figure}

We formulate the problem of selecting $M$ out of $N$ state-transitioning arms at each time step, as an RMAB, with beneficiaries being the arms in preventive healthcare intervention scenarios. We represent the engagement pattern as an MDP (Figure~\ref{fig:MDP}) with three (abstract) ``behavioral'' state: %Each beneficiary is in one of the three states and gradually moves from one state to another: \textit{Self Motivated} to \textit{Persuadable} to \textit{Lost Cause} if no intervention is provided. Each behavioral state is an abstraction of engagement values.\\
(i)~\textit{self motivated} ($S$): in this state, beneficiary shows high engagement and there is no need to intervene in this state. (ii)~\textit{persuadable} ($P$): in this state, the beneficiary engages less frequently with a possibility of increasing engagement when intervened, which makes this the most interesting state, and 
%. Thus, it is important to identify beneficiaries who have reached this state.\\
(iii)~\textit{lost Cause} ($L$): The engagement is very low in this state and very likely to remain low irrespective of intervention. 

These three states capture different levels of engagement as well as differences in terms of the benefit obtained by interventions, which is not possible to represent using other existing intervention models, such as the model described by Mate et al.~\citeyear{Mate2020}. Note that the more the states, slower is the convergence of any online algorithm. Thus, for short-term intervention programs (a year or two), we recommend a three-states model. In Section~\ref{sec:case-study}, we provide a mechanism to obtain the states from real-data. Let $A_i(t)\in\{0,1\}$ be the action taken on beneficiary $i$ at time $t$; $1$ denotes an intervention and $0$ otherwise. Also, $\sum_i A_i(t) = M$ for each time slot $t$. Depending on the action taken, each beneficiary changes its states according to the transition probabilities. 

% \begin{enumerate}
%     \item \textbf{Self Motivated} ($S$): In this state, beneficiary picks up most of the calls and listens to them almost completely, thus, showing high engagement. Naturally, there is not much necessity of providing interventions in this state.
%     \item \textbf{Persuadable} ($P$): In this state, the beneficiary engages less frequently. However, there is a possibility of increasing engagement when an intervention is given to them. Thus, it is important to identify beneficiaries who have reached this state, so that proper interventions can be provided to them. 
%     \item \textbf{Lost Cause} ($L$): Typically, the extent of engagement is very low in this state, and there is a high chance that the engagement remains low regardless of any intervention. Once a beneficiary reaches this state, they are likely to remain disengaged in near future. 
% \end{enumerate}

%Each behavioral state is an abstraction of engagement values.
%---for example, listening to more than $50\%$ of the call can be considered as being in state $S$. 
%Similarly, listening to $5-50\%$ of the call is considered as being in state $P$, and listening to $0-5\%$ of the call is equivalent to being in state $L$. 

Though our model is inspired by a preventive intervention in healthcare, this formulation captures the intervention problem in many other domains, such as sensor monitoring, anti-poaching patrols, and uplift modeling in eCommerce. Depending on the domain, a persuadable state is where the benefit of an intervention is maximum. When a beneficiary is in the persuadable state and intervention is provided, an arm is more likely to transition to the self-motivated state whereas, when no intervention is provided, it is more likely to transition to the lost-cause state. These transitions are denoted as $p^{PS}_{i,1}$ and $p^{PL}_{i,0}$ for action $1$ and $0$, respectively. %Note that the outgoing transition probabilities from states $S$ and $L$ remain the same irrespective of the action, and differs only for the state $P$. 
Every transition generates a state-dependent reward. Since better engagements have higher rewards, we assume $R_i^S> R_i^P>R_i^L$. The transition probabilities $p_{i,a}^{YZ}$ may be different for each beneficiary and thus, the effect of intervention would vary from one beneficiary to another. 
%Hence, it is important to select $M$ beneficiaries so as to maximize the positive effect of the intervention.
If the transition probabilities were known beforehand, one could compute the Whittle-index policy and select accordingly. However, here, transition probabilities are unknown, and hence, we consider the problem of learning the Whittle index while simultaneously selecting a set of best arms depending on the estimated Whittle Index.

\section{Whittle Index based Q-learning (WIQL)}
Q-Learning~\cite{watkins1992q} is a well-studied reinforcement learning algorithm for estimating $Q^*(Z,a)$ for each state-action pair ($Z,a$) of an MDP. 
\[Q^*(Z,a):= R^Z + \displaystyle \sum_{Z'\in\{S,P,L\}} \mathcal{P}(Z,a,Z')\cdot V^*(Z'),\]
where the optimal expected value of a state is given by 
\[V^*(Z):= \underset{a\in\{0,1\}}{\max} \left(R^Z + \displaystyle \sum_{Z'\in\{S,P,L\}} \mathcal{P}(Z,a,Z')\cdot V^*(Z')\right)\]

Q-Learning estimates $Q^*$ using point samples---at each time $t$, an agent (policy maker) takes an action $a$ using estimated $Q$ values at the current state $Z$, a reward $R$ is observed, a new state $Z$ is reached, and $Q$ values are updated according to the following update rule:
    \begin{eqnarray}
    Q^{t+1}(Z,a)&\leftarrow & (1-\alpha_t(Z,a))\cdot Q^t(Z,a) +\nonumber \\
    &&\hspace{-0.2in} \alpha_t(Z,a)\!\cdot\!\left(R^Z(t) + \underset{a'\in\{0,1\}}{\max} Q^t(Z',a')\right)\ \label{eq:Q-update}
    \end{eqnarray}
Here, $\alpha_t(\cdot)\in[0,1]$ is the learning parameter. When $\alpha=0$, the agent does not learn anything new and retains the value obtained at $t^{th}$ step, while $\alpha=1$ stores only the most recent information and overwrites all previously obtained
rewards. Setting $0 < \alpha < 1$ strikes a balance between the new values and the old ones. With mild assumptions on $\alpha$, the convergence of Q-Learning to the optimal $Q^*$ values has been established~\cite{watkins1992q,jaakkola1994convergence,borkar2000ode}. We build on the results to provide a Q-Learning approach for learning Whittle-index policy.

We adopt Q-Learning for RMABs and store Q values separately for each arm. Q values of state-action pairs are typically used for selecting the best action for an arm at each state; however, for RMAB, the problem is to select $M$ arms. Our action-selection method ensures that an arm whose estimated benefit is higher is more likely to get selected. Algorithm~\ref{algo:QL} describes the action-selection and the update rule. This algorithm is \textit{not} specific to any particular state representation and can be used for any finite state RMAB instance. However, when the number of states are large (10 states), the convergence is typically slow and not suitable for short horizon.

\begin{algorithm}[!t]
\DontPrintSemicolon
{\small
  \KwInput{$N$, $M$, $\alpha(c)$, and initial states $X_i(0)\in\mathcal{S}\  \forall\ i\in [N]$.}
  {\textbf{Initialize:} $Q_i^0(Z,a)\leftarrow 0$ and $\lambda_i^0(Z)\leftarrow 0$ for each state $Z\in\mathcal{S}_i$, each action $a\in\{0,1\}$,  each arm $i\in[N]$}.
  
  %\tcp*{Q values are initialized to $0$.}
  %\textbf{Initialize:} $\lambda_i^0(Z)\leftarrow 0$ for each state $Z\in\mathcal{S}_i$ and each arm $i\in\{1,\ldots N\}$.
  %\tcp*{Whittle indices are initialized to $0$.}
  \For{$t$ in $1,\ldots,T$}
  {
      \tcp{Select $M$ arms using $\epsilon$-decay policy}
      {$\epsilon \leftarrow \frac{N}{N+t}$}.
      
      {With probability $\epsilon$, select $M$ arms uniformly at random. Otherwise, select top $M$ arms according to their values $\lambda_i^{t}(X_i(t))$.  Store the selected arms in $\Psi$.}
      
      \tcp{Take suitable actions on the arms}
      \For{ $i$ in $1:N$}
      {
          \If{$i\in\Psi$}
            {
                {Take action $A_i(t)=1$ on arm $i$}.
            }
            \Else
            {
            	{Take action $A_i(t)=0$ on arm $i$}.
            }

           {Observe reward $r$ and next state $X_i(t+1).$}
        }
        
        \tcp{\small Update Q, $\lambda$ using $Z=X_i(t)$, $a=A_i(t)$} {$c^h_{i,Z,a} =\displaystyle\sum_{h=0}^t \mathbb{I}\{X_i(h)=Z \mbox{ \& } A_i(h) = a\}$.}
        
        \For{ $i$ in $1:N$} 
         {
            {$Q_i^{t+1}(Z, a) \leftarrow (1- \alpha(c^h_{i,Z,a}))\cdot Q_i^{t}(Z, a)+ \alpha(c^h_{i,Z,a}) \cdot \left(r + \underset{a'\in\{0,1\}}{\max} Q_i^{t}(X_i(t+1),a')\right)$.}

            {$\lambda_i^{t+1}(X_i(t)=Z)\leftarrow Q_i^{t+1}(Z, 1) - Q_i^{t+1}(Z, 0)$}
        }
    }

\caption{Whittle Index based Q-learning(WIQL)}
\label{algo:QL}
}
\end{algorithm}

We propose Whittle Index based Q-learning (WIQL), that uses an $\epsilon$-decay policy to select $M$ arms at each time step $t$. During early steps, arms are likely to be selected uniformly at random. As time proceeds, arms with higher values of their estimated $\lambda_i(X_i(t))$ gets more priority. The selected set of $M$ arms (who receive interventions) is called $\Psi$. Each arm is \textit{restless}, i.e., each arm transitions to a new state and observes a reward, with or without interventions. These observations are then used for updating the Q values in Step $13$ (Eq.~\ref{eq:Q-update}). While updating $Q_i(Z,a)$, we use a learning parameter $\alpha(c_{i,Z,a})$ that decreases with increase in $c^t_{i,Z,a}$ (number of times the arm $i$ received action $a$ at state $Z$); eg, $\alpha(c^t_{i,Z,a})=1/(c^t_{i,Z,a}+1)$ satisfies this criteria. These Q values are then used to estimate the Whittle index $\lambda(X_i(t))$. 

\section{Theoretical Results}
In this section, we show that the WIQL does not alter the optimality guarantees of Q-learning. First, we show that taking intervention decisions based on the benefit, (i.e., difference in Q values of active and passive actions) is equivalent to optimizing joint Q over all arms subject to the budget on intervention actions. 

\begin{theorem}
Taking action $1$ on the top $M$ arms according to $(Q^*_i(s_i, 1) - Q^*_i(s_i, 0))$ is equivalent to maximizing $\sum_i Q_i^*(s_i, a_i)$ over all possible action profiles $(a_1,\ldots,a_N)\in\{0,1\}^N$ such that $\sum_i a_i = M$.
\label{th:1}
\end{theorem}
\textit{Proof Sketch.} For ease of explanation, we prove this for M=1. Let $i^*$ be the  arm that maximizes the benefit of taking an intervention action ($Q^*_{i^*}(s_{i^*}, 1) - Q^*_{i^*}(s_{i^*}, 0)$) at its current state $s_{i^*}$. Then, for any $j \in\{1,\ldots, N\}\setminus\{i^*\}$.
\begin{align}
    &Q^*_{i^*}(s_{i^*}, 1) - Q^*_{i^*}(s_{i^*}, 0) \geq Q^*_{j}(s_{j}, 1) - Q^*_{j}(s_{j},0) \nonumber\\
    &Q^*_{i^*}(s_{i^*}, 1) + Q^*_{j}(s_{j},0) \geq Q^*_{j}(s_{j}, 1) + Q^*_{i^*}(s_{i^*}, 0)\label{eq:i*} \\
    &\text{Adding $\sum_{i \neq i*, i \neq j} Q_i^*(s_i,0)$ on both sides} \nonumber\\
    &Q^*_{i^*}(s_{i^*}, 1)\!\! +\!\! \displaystyle \sum_{i\neq i^*} Q^*_{i}(s_{i} 0) \geq Q^*_{j}(s_{j}, 1)\!\! +\!\! \displaystyle \sum_{i\neq j} Q^*_{i}(s_{i} 0).\label{eq:max}
\end{align}
Eq.~(\ref{eq:max}) shows that taking intervention action on $i^*$ and passive actions on all other arms would maximize $\sum_i Q^*(s_i, a_i)$ when $M$=$1$. This argument holds true for any $M\geq 1$ (complete proof is in the Appendix).$\hfill \square$
\iffalse
, let $I^{*}$ be the set that contains the top k beneficiaries (i.e., beneficiaries with highest values of $Q_i(s_i, 1) - Q_i(s_i, 0)$), and $I^{-*}$ be the set contains the non top k beneficiaries. Let $I_k$ represent any set of k beneficiaries. We then have:
\begin{align}
&\sum_{i^* \in I^{*}} [Q^*_{i^*}(s_{i^*}, 1) - Q^*_{i^*}(s_{i^*}, 0)] \geq \sum_{j \in I_k} [Q^*_{j}(s_{j}, 1) - Q^*_{j}(s_{j},0)] \nonumber\\
&\text{Adding $\sum_{i \notin I^*, i \notin I_k} Q_i^*(s_i,0)$ on both sides, we have} \nonumber\\
&\sum_{i^* \in I^{*}} Q_{i*}^{*}(s_{i^*},1) + \sum_{i \in I^{-*}} Q_{i}^*(s_i,0) \geq \sum_{i \in I_k} Q_i(s_i,1) + \sum_{i \notin I_k} Q_i(s_i,0) \label{eq:maxk}
\end{align}
Equation~\ref{eq:maxk} shows that taking intervention action on beneficiaries in $I^*$ and passive actions on all other arms would maximize $\sum_i Q^*(s_i, a_i)$ when M = k. 
\fi 

\begin{theorem}
WIQL converges to the optimal with probability $1$ when $\sum_h\alpha(c^h_{i,Z,a}) = \infty \mbox{ and } \sum_h \alpha(c^h_{i,Z,a})^2 < \infty.$
\end{theorem}
\textit{Proof Sketch.} This proof follows from (1)~the convergence guarantee of Q-Learning algorithm~\cite{watkins1992q}, (2)~$\epsilon$-decay selection policy, and (3)~theorem~\ref{th:1}. It has been established in \cite{watkins1992q} that the update rule of Q-Learning converges to $Q^*$ whenever $\sum_t\alpha_t(Z,a) = \infty$  and $ \sum_t \alpha_t(Z,a)^2 < \infty.$ These assumptions require that all state-action pairs be visited infinitely often, which is guaranteed by the $\epsilon$-decay selection process, where each arm has a non-zero probability of being selected uniformly at random. Thus, $Q^t_i(Z,a)$ converges to $Q^*_i(Z,a)$ which implies that $Q^t_i(Z,1) - Q^t_i(Z,0)$ converges to $Q^*_i(Z,1)-Q^*_i(Z,0)$ (using series convergence operation). Finally, using Theorem~\ref{th:1}, we claim that selecting arms based on highest values of $Q^*_i(Z,1)-Q^*_i(Z,0)$ would lead to an optimal solution problem. This completes the proof that WIQL converges to the optimal. $\hfill \square$

%Next, we evaluate WIQL on several well-studied numerical examples and compare its performance with existing learning algorithms for Whittle Indices. 

\section{Experimental Evaluation}
We compare WIQL against five benchmarks: (1)~\textbf{OPT}: assumes full knowledge of the underlying transition probabilities and has access to the optimal Whittle Indices, (2)~\textbf{AB}~\cite{avrachenkov2020whittle}, (3)~\textbf{Fu}~\cite{fu2019towards}, (4)~\textbf{Greedy}: greedily chooses the top $M$ arms with the highest difference in their observed average rewards between actions $1$ and $0$ at their current states, and (5)~\textbf{Random}: chooses $M$ arms uniformly at random at each step.

We consider a numerical example and a maternal healthcare application to simulate RMAB instances using beneficiaries' behavioral pattern from the call-based program. 
%for maternal healthcare (detailed in Section~\ref{sec:case-study}). %While executing WIQL, we consider a fixed value of learning rate $\alpha=0.1$.
For each experiment, we plot the total reward averaged over $30$ trials, to reduce the effect of randomness in the action-selection policy.

\subsection{Numerical Example: Circulant Dynamics}\label{sec:numerical}
This example has been studied in the existing literature on learning Whittle Indices~\cite{avrachenkov2020whittle,fu2019towards}. Each arm has four states $\mathcal{S} = \{0,1,2,3\}$ two actions $\mathcal{A}=\{0,1\}$. The rewards are $R^0=-1$, $R^1=R^2=0$, and $R^3=1$ for $a\in\{0,1\}$. The transition probabilities for each action $a\in\{0,1\}$ are represented as a $|\mathcal{S}|\times |\mathcal{S}|$ matrix:
\begin{equation*}
\mathcal{P}^1\!\! = \!\!
\begin{pmatrix}
0.5 & 0.5 & 0 & 0 \\
0 & 0.5 & 0.5 & 0 \\
0 & 0 & 0.5 & 0.5  \\
0.5 & 0 & 0 & 0.5 
\end{pmatrix}, \mathcal{P}^0 \!\!= \!\!
\begin{pmatrix}
0.5 & 0 & 0 & 0.5 \\
0.5 & 0.5 & 0 & 0 \\
0 & 0.5 & 0.5 & 0 \\
0 & 0 & 0.5 & 0.5 
\end{pmatrix}
\end{equation*}
The optimal Whittle Indices for each state are as follows: $\lambda^*(0)=-0.5$, $\lambda^*(1) = 0.5$, $\lambda^*(2)=1$, and $\lambda^*(3)=-1$. Whittle Index policy would prefer taking action $1$ on arms who are currently at state $2$, followed by those at state $1$, then those at state $0$, and lastly those at state $3$.

\begin{figure}[t]
\centering
\begin{subfigure}{0.99\columnwidth}
  \centering
  \includegraphics[width=.95\linewidth]{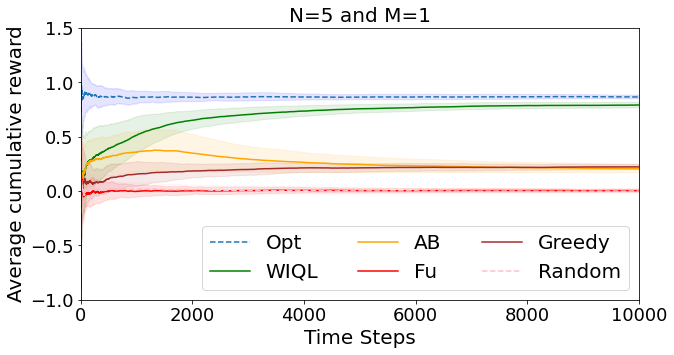}
  \caption{Average reward: $1$ out of $5$ arms chosen.}
  \label{fig:e1n5m1}
\end{subfigure}%
\vfill
\begin{subfigure}{0.99\columnwidth}
  \centering
  \includegraphics[width=.95\linewidth]{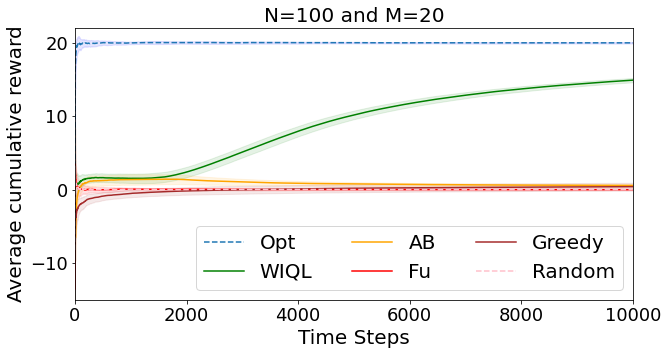}
  \caption{Average reward: $20$ out of $100$ arms chosen.}
  \label{fig:e1n100m20}
\end{subfigure}
\caption{Circulant Dynamics: Results averaged over $30$ trials.}
\label{fig:e1}
\end{figure}

Figure~\ref{fig:e1} demonstrates that WIQL gradually increases towards the OPT policy, for two sets of experiments---(1) $N$=$5$ and $M$=$1$ and (2) $N$=$100$ and $M$=$20$. We observe that AB converges towards an average reward of zero. This happens because it prioritizes arms who are currently at state $1$ over other states. Since, the expected reward of taking action $1$ at state $1$ is $\mathcal{P}(1,1,1)\cdot R^1 + \mathcal{P}(1,1,2)\cdot R^2 = 0$, the total average reward also tends to zero. The result obtained by the algorithm Fu is the same as what is shown in Figure $2$ of their paper~\cite{fu2019towards} where the total average reward converges to a value of $0.08$. As expected, Greedy and Random, one being too myopic and the other being too exploratory in nature, are unable to converge to the optimal value.

These observations show that WIQL outperforms the existing algorithms on the example that was considered in the earlier papers. Note that, while implementing the algorithms AB and Fu, we fixed the hyperparameters to the values specified for this example. However for the real-world application, that we consider next, it is not obvious how to obtain the best set of hyperparameters for their algorithms. Thus, we do not compare these algorithms for the maternal healthcare application. Next, we compare the performance of WIQL algorithm with Greedy, Random and a Myopic policy (defined in the subsequent paragraph).

\subsection{Real-world Application: Maternal Healthcare}\label{sec:case-study}

\begin{figure*}[!h]
\centering
\begin{subfigure}{0.66\columnwidth}
  \centering
  \includegraphics[width=.99\linewidth]{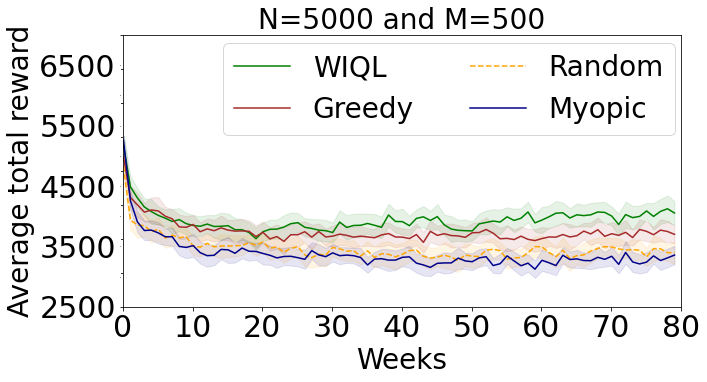}
  %\caption{Average reward accumulated when $M=1$.}
  %\label{fig:e2n5m1}
\end{subfigure}%
\begin{subfigure}{0.66\columnwidth}
  \centering
  \includegraphics[width=.99\linewidth]{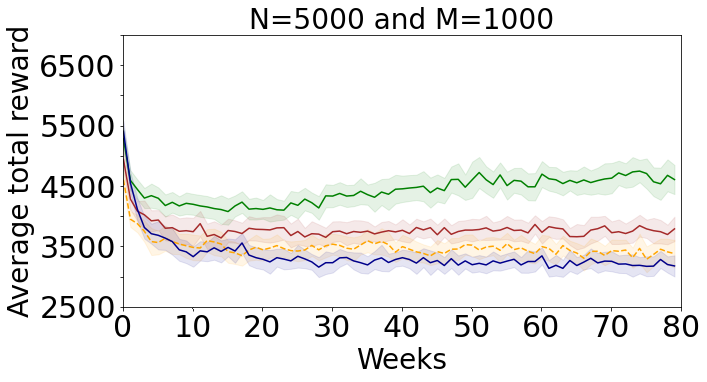}
\end{subfigure}
\centering
\begin{subfigure}{0.66\columnwidth}
  \centering
  \includegraphics[width=.99\linewidth]{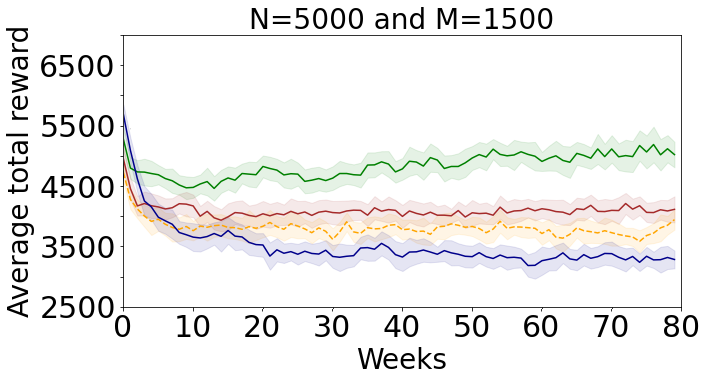}
\end{subfigure}%
\vfill
\begin{subfigure}{0.66\columnwidth}
  \centering
  \includegraphics[width=.99\linewidth]{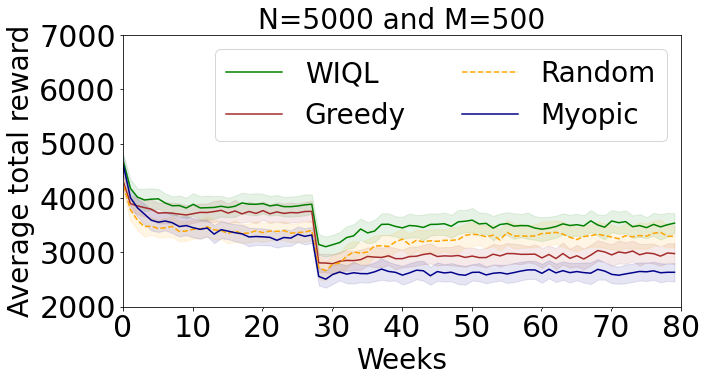}
\end{subfigure}
\centering
\begin{subfigure}{0.66\columnwidth}
  \centering
  \includegraphics[width=.99\linewidth]{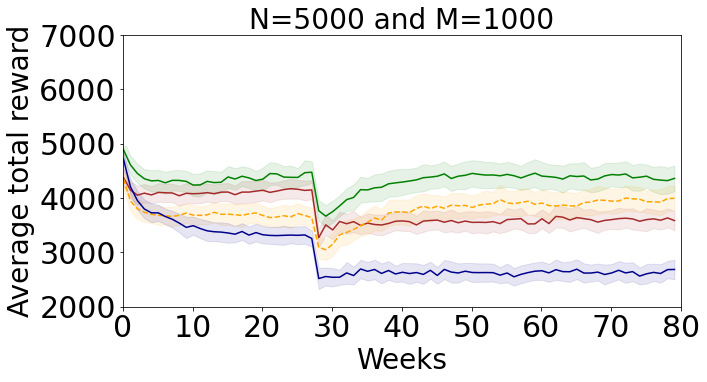}
\end{subfigure}%
\begin{subfigure}{0.66\columnwidth}
  \centering
  \includegraphics[width=.99\linewidth]{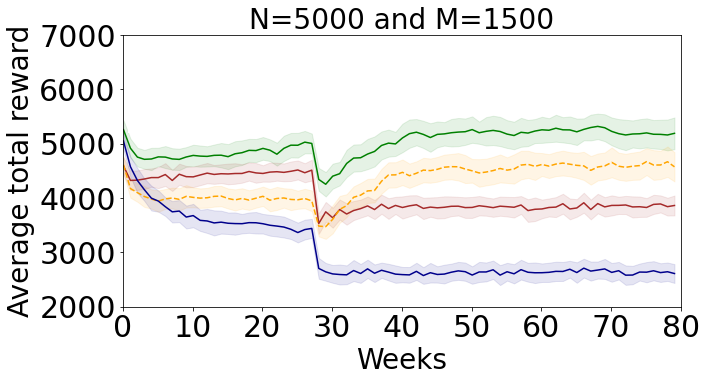}
\end{subfigure}
\caption{Evaluating WIQL on maternal healthcare application with static and dynamic behavior change (shown in first and second row respectively). The graphs show the total reward accumulated on a particular week, averaged over 30 iterations.}
\label{fig:e3}
\end{figure*}

We now focus on the \textit{maternal healthcare intervention problem} 
%(detailed in Section~\ref{sec:model}) 
where only a small subset of beneficiaries can be selected for providing interventions every week. 
We use the data, obtained from the call-based preventive care program, that contains call-records of enrolled beneficiaries---how long they listened to the call, whether an intervention was given, and when. The data also contain the ID of a healthcare worker dedicated to provide personalized intervention to each enrolled beneficiary. 
The data was collected for an experimental study towards building up a robust intervention program (which is the focus of this work). During the experimental study, only one intervention (in-person visit by a health worker) was provided to a selected set of beneficiaries ($1559$ chosen from among $3031$ beneficiaries) who were more likely to drop-out of the program.
%\footnote{Out of a total of $4273$ beneficiaries, $1242$ of them did not receive the automated calls sent to them (we speculate that the current COVID-19 pandemic is one of the reasons for the low engagement). So, we remove them from the dataset.}).  
We call this the \textbf{Myopic} intervention and use it as a benchmark to compare our approach.

We use the three-state MDP model (Figure~\ref{fig:MDP}) to simulate the behavior of the beneficiaries. During a particular week, the beneficiaries listening to $>50\%$ of the content of the automated calls are in state $S$ (self-motivated), those listening to $5-50\%$ are in state $P$ (persuadable), and those listening to $<5\%$ are in state $L$ (lost cause). 
A reward of $2$ is obtained when a beneficiary is in state $S$, a reward of $1$ is obtained in state $P$, and a reward of $0$ is obtained in state $L$. 
Thus, a high total reward accumulated per week implies that a large number of beneficiaries are at either state $S$ or $P$.

Some other observations on the call-records are as follows. The beneficiaries who were in state $P$ on a particular week never transitioned to state $S$ in the immediate next week, unless they received an intervention. On the other hand, a few beneficiaries who were in state $L$ transitioned to state $P$ even without intervention. Moreover, the fraction of times a transition from state $L$ to $P$ occurred is almost the same with and without the intervention at state $L$. %This behavior can be modeled as an MDP described in Figure~\ref{fig:MDP} with suitable state transition probabilities. 
Even though these transition probabilities at the level of all users are known based on this data, it is difficult to know the actual transitions for a given beneficiary {\em a priori} and hence, we simulate this behavior. We conduct two set of simulations, namely \textit{static} (transition probabilities of each beneficiary remain same throughout their enrollment), and \textit{dynamic} (transition probabilities change a few weeks after their enrollment).

\paragraph{Static State-Transition Model.}%\subsubsection*{}\label{sec:static}
We assume three categories of arms---(A)~high chance of improvement: highly likely to improve their engagement on receiving an intervention, and deteriorate in absence of an intervention when they are at state $P$, i.e., $p^{PS}=0.8$ and $p^{PL}=0.8$, (B)~medium chance of improvement: $p^{PS}=0.4$ and $p^{PL}=0.6$, and (C)~low change improvement: $p^{PS}=0.1$ and $p^{PL}=0.6$. We assume $10$ arms belong to category-A, $10$ arms belong to category-B and $30$ arms belong to category-C. This assumption helps us determine the efficacy of any learning algorithm; in particular, the most efficient algorithm would quickly learn to intervene the $10$ arms of category-A whenever they are at state $P$. We compare WIQL with greedy, random, and \textit{Myopic} algorithm. %As mentioned earlier, the Myopic algorithm had been a standard way of selecting beneficiaries for intervention---select the beneficiaries with the lowest rate of engagement. 

\paragraph{Dynamic State-Transition Model.}
For the dynamic setting, we simulate the first $28$ weeks as described in Section~7.2.1. Further, we assume that the arms in category (A) change to ``medium'' chance of improvement after they are enrolled for around $28$ weeks, those in category (B) change to ``low'' chance of improvement. Also, there is a new set of $10$ arms with ``high'' chance of improvement. Note that, in reality, the behavior change would happen at any arbitrary time; however, to check if WIQL adapts to dynamic change of transition probabilities, we set this simulation environment.

We run each simulation for $80$ weeks (the length of the program). 
%(since a beneficiary is enrolled for about $80$ weeks during her pregnancy, childbirth, and up to one year after the childbirth). 
In Figure~\ref{fig:e3} we provide results obtained by considering various values of $M\in\{500,1000,1500\}$, where a value of $M$ represents the total number of personalized visits made by the $100$ health-care workers on a week. We observe that, for $M=500$ (each health-care worker visits only $5$ beneficiaries per week), WIQL performs (only) marginally better than the other benchmarks. %In fact, all the algorithms achieve an average engagement of around $3500$, which implies that at least $1500$ beneficiaries are in the state $L$ (where the reward is $0$) by the end of $80$ weeks. Therefore, with a very low value of $M$, a majority of beneficiaries would end up in state $L$ and this would negatively impact the success of the call-based program. 
In contrast, when $M\in\{1000,1500\}$, the reward obtained by WIQL is higher than Greedy and significantly more than Myopic and Random. Comparing WIQL and Greedy based on their total reward per week, we observe that the per-week engagement outperforms Greedy by a significant margin. Observe that, the convergence of WIQL to a total reward of $5000$ is quicker when $M$ is higher. This is because more sample points are observed per week. Additionally, we observe that the \textit{myopic} algorithm leads to overall low engagement among the beneficiaries, even under the static setting. %Thus, we conclude that providing interventions only to the low-engagement beneficiaries may lead to poor overall benefit. 
These results show that WIQL is able to learn which arms should be intervened at which state without any prior knowledge about the transition probabilities, and also adapts better to the underlying transition dynamics. 

\section{Conclusion and Discussion}
We focus on a limited-resource sequential decision problem and formulate it as an RMAB setting. We provide a mechanism to systematically learn and decide on whom to intervene, and hence, improve the overall benefit of intervention. Our method possesses the capacity of impacting and improving the well-being of millions, for example, in the maternal healthcare domain, as demonstrated in this paper.
%Such an approach is not only applicable in health interventions, but also for problems related to monitoring tasks like sensor/machine maintenance~\cite{Ianello2012,Glazebrook2006} and anti-poaching patrols~\cite{Qian2016}, or uplift modeling in eCommerce platforms~\cite{gubela2019conversion}. 
%Due to the absence of prior information, we can not directly compute Whittle Indices for the RMAB problem. We propose a method, called WIQL, which dynamically learns the Whittle Index policy. We show that WIQL converges to Whittle Indices under mild assumptions.
%We then empirically compare WIQL on two examples that are considered in the earlier papers. We observe that WIQL outperforms the benchmarks in terms of the average accumulated reward. We then simulate the maternal health intervention problem using the observations of the data obtained by the call-based health intervention program and show that WIQL eventually learns to select beneficiaries who are more likely to improve their engagement in getting an intervention. 

WIQL is a general solution for learning RMABs, and we demonstrate this using examples from other domains, such as Circulant Dynamics. Additionally, WIQL can be used in a more general setting where new arms arrive over time. For the departing arms, we can assume that each arm ends in a new state ``dropped'' and never transitions to the earlier states. Moreover, our experiments show that WIQL adapts to the dynamic behavioral change. In practice, however, WIQL may not be directly applicable for domains where there are beneficiaries with extreme health risk. The non-zero probability of being selected randomly may come at a cost of a patient in critical need of intervention. One way to mitigate this issue is to assign priorities to beneficiaries depending on their comorbidities, possible complication during pregnancy and after childbirth, etc. If the high-risk class is small, we can target the intervention for all of them, and run WIQL on the remaining beneficiaries. This constraint may hamper the convergence guarantee of WIQL; however, it would benefit the enrolled women at large. 

Going ahead, it would be interesting to solve the RMAB learning problem considering a large number of states and more than two actions, each with a different cost of operation.

%% The file named.bst is a bibliography style file for BibTeX 0.99c
\bibliographystyle{named}
\bibliography{references}
\appendix

\section{Proof for Whittle Index Q-Learning}
\begin{theorem}
Taking action $1$ on the top $M$ arms according to $(Q^*_i(s_i, 1) - Q^*_i(s_i, 0))$ is equivalent to maximizing $\sum_i Q_i^*(s_i, a_i)$ over all possible action profiles $(a_1,\ldots,a_N)\in\{0,1\}^N$ such that $\sum_i a_i = M$.
\end{theorem}
\textbf{Proof} Let $I^{*}$ be the set containing the top $M$ beneficiaries (i.e., beneficiaries with highest values of $Q_i(s_i, 1) - Q_i(s_i, 0)$), and $I^{-*}$ be the set contains the non top M beneficiaries. Let $I_M$ represent any set of k beneficiaries. We then have:
\begin{align}
&\sum_{i^* \in I^{*}} [Q^*_{i^*}(s_{i^*}, 1) - Q^*_{i^*}(s_{i^*}, 0)]\nonumber\\&\quad\qquad\qquad \geq \sum_{j \in I_M} [Q^*_{j}(s_{j}, 1) - Q^*_{j}(s_{j},0)] \nonumber\\\\
&\text{Adding $\sum_{i \notin I^*, i \notin I_M} Q_i^*(s_i,0)$ on both sides, we obtain:} \nonumber\\
&\sum_{i^* \in I^{*}} Q_{i*}^{*}(s_{i^*},1) + \sum_{i \in I^{-*}} Q_{i}^*(s_i,0)\nonumber\\
&\quad\qquad\qquad \geq \sum_{i \in I_M} Q_i(s_i,1) + \sum_{i \notin I_M} Q_i(s_i,0) \label{eq:maxk}
\end{align}
Equation~\ref{eq:maxk} shows that providing intervention to beneficiaries in $I^*$ and taking passive actions on all other arms would maximize $\sum_i Q^*(s_i, a_i)$. 

Now the rest of the proof follows from (1)~the convergence guarantee of Q-Learning algorithm~\cite{watkins1992q}, (2)~$\epsilon$-decay selection policy, and (3)~the indexability assumption of RMABs~\cite{whittle1988restless}. It has been established in \cite{watkins1992q} that the update rule of Q-Learning converges to $Q^*$ whenever $\sum_t\alpha_t(Z,a) = \infty$  and $ \sum_t \alpha_t(Z,a)^2 < \infty.$ These assumptions require that all state-action pairs be visited infinitely often, that is, the underlying MDP do not have a state $Z$ such that after some finite time $t$, $Z$ is not reachable via any sequence of actions. Similarly, the assumptions $\displaystyle\sum_h\alpha(c^h_{i,Z,a}) = \infty \mbox{ and } \sum_h \alpha(c^h_{i,Z,a})^2 < \infty$ require that each arm, state and action pair would be visited infinitely often. Thus, in addition to the reachability assumption on MDP, we need to show that each arm is visited infinitely often. This is guaranteed by the $\epsilon$-decay selection process, where each arm has a non-zero probability of being selected uniformly at random. Thus, $Q^t_i(Z,a)$ converges to $Q^*_i(Z,a)$ which implies that $Q^t_i(Z,1) - Q^t_i(Z,0)$ converges to $Q^*_i(Z,1)-Q^*_i(Z,0)$ (using series convergence operation). Note that the value $Q^*_i(Z,1)-Q^*_i(Z,0)$ is the Whittle Index. Using these values for selecting the best arms would lead to an optimal solution for the relaxed Lagrangian problem, because of the indexability assumption (Definition~\ref{def:indexability}). $\hfill \square$

%Next, we evaluate WIQL on several well-studied numerical examples and compare its performance with existing learning algorithms for Whittle Indices. 

\section{Experimental Evaluation}

Here we provide the details of the benchmark algorithms that we use for evaluation. Also, we provide examples from two other domains, one of them was considered in~\cite{avrachenkov2020whittle} and the other was considered in~\cite{fu2019towards}.

\subsection{Benchmark Algorithms}
We compare WIQL with five other algorithms:
\begin{itemize}
    \item \textbf{OPT}: This algorithm assumes full knowledge of the underlying transition probabilities and has access to the optimal Whittle Indices $\lambda_i^*(Z)$ for each arm $i$ and each state $Z$. At each time step $t$, the top $M$ arms are chosen according to the $\lambda^*$ values of their current states. OPT is our benchmark to visualize the convergence of other algorithms.
    \item \textbf{AB}: Proposed by~\cite{avrachenkov2020whittle} this method is based on a fundamental change in the definition of Q values that aims at converging to the optimal Whittle Index policy. Arms are assumed to be homogeneous and they use a shared Q update step for all the arms. The shared update helps in collecting the point samples of Q values very quickly and results in fast convergence to the optimal algorithm. However, we consider scenarios where the arms may have different transition probabilities and we need to differentiate among arms, even when they are in the same state. Thus, for our experiments, we store Q values separately for each arm.
    \item \textbf{Fu}: This method has been proposed by~\cite{fu2019towards} to update Q values separately for each arm, each $\lambda\in\Lambda$, and each state-action pair. They assume $\Lambda$ to be a set of input parameters. At each time step, $\lambda_{min}\in \Lambda$ is computed which represents the subsidy that minimizes the gap between Q values for taking action $1$ and that of action $0$. For the experiments, we set $\Lambda$ and other hyperparameters the same as what they considered in their paper. 
    This method is a heuristic with no convergence guarantee. 
    \item \textbf{Greedy}: This method maintains a running average reward for each arm, state, and action, depending on the history of observations. At each time step, it greedily chooses the top $M$ arms with the highest difference in their average rewards between actions $1$ and $0$ at their current states. 
    \item \textbf{Random}: This method chooses $M$ arms uniformly at random at each step. Note that this algorithm is the same as setting $\epsilon=1$ in Step $4$ of the WIQL algorithm. 
\end{itemize}

\subsection{Numerical Examples}
In this section, we provide two additional numerical examples for more general RMAB instances. 
\subsubsection{Example with restart}
This RMAB problem is considered by~\cite{avrachenkov2020whittle}. Here, each arm is assumed to be in one of the five states $\mathcal{S} = \{0,\ldots, 4\}$ at any point of time. There are two actions---active ($a=1$) and passive ($a=0$). The active action forces
an arm to restart from the first state. At each time step, action $a=1$ can be taken only on $M$ arms. The reward from an arm $i$ at a state $Z\in\mathcal{S}$ is assumed to be $R^Z_i=0.9^{Z}$ when passive action is taken. The transition probabilities for each action $a\in\{0,1\}$ are represented as a $|\mathcal{S}|\times |\mathcal{S}|$ matrix, where $p^1= 1.0$, $q^1=0.0$, $p^0= 0.1$, and $q^0=0.9$. 
\begin{equation*}
\mathcal{P}^a = 
\begin{pmatrix}
p^a & q^a & 0 & 0 & 0 \\
p^a & 0 & q^a & 0 & 0 \\
p^a & 0 & 0 & q^a  & 0 \\
p^a & 0 & 0 & 0 & q^a\\
p^a & 0 & 0 & 0 & q^a 
\end{pmatrix}
\end{equation*}.

\begin{figure*}[!h]
\centering
\begin{subfigure}{0.9\columnwidth}
  \centering
 \includegraphics[width=\linewidth]{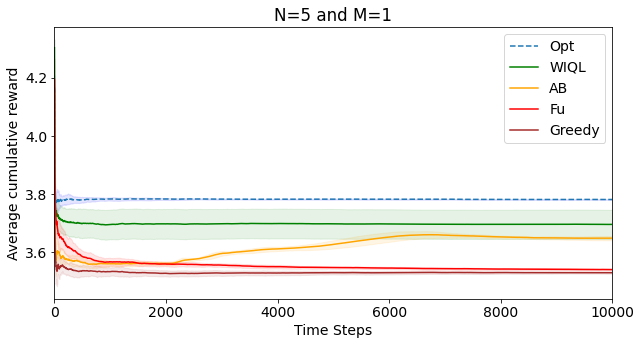}
  \caption{Average reward accumulated when only $1$ arm out of $5$ arms can be chosen at each iteration.}
  \label{fig:e4n5m1}
\end{subfigure}%
\hfill
\begin{subfigure}{0.9\columnwidth}
  \centering
  \includegraphics[width=\linewidth]{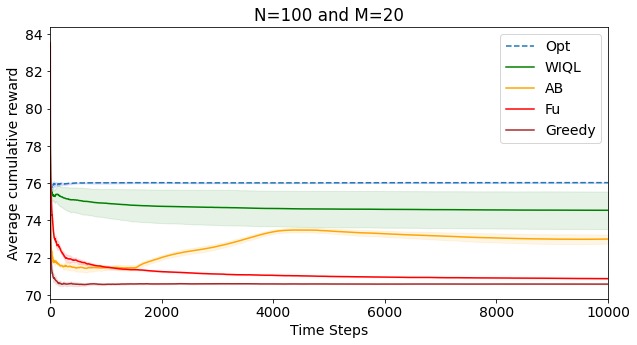}
  \caption{Average reward accumulated when only $20$ arms out of $100$ arms can be chosen at each iteration.}
  \label{fig:e4n100m20}
\end{subfigure}
\caption{Comparing various algorithms for the example with Restart. The results are averaged over $30$ trials.}
\label{fig:e4}
\end{figure*}

\begin{figure*}[!h]
\centering
\begin{subfigure}{0.9\columnwidth}
  \centering
  \includegraphics[width=.9\linewidth]{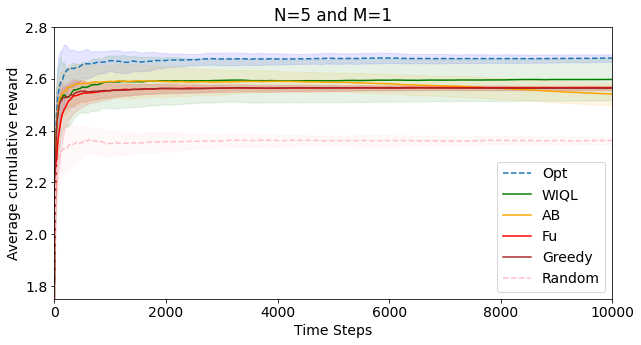}
  \caption{Average reward accumulated when only $1$ arm out of $5$ arms can be chosen at each iteration.}
  \label{fig:e2n5m1}
\end{subfigure}%
\hfill
\begin{subfigure}{0.9\columnwidth}
  \centering
  \includegraphics[width=.9\linewidth]{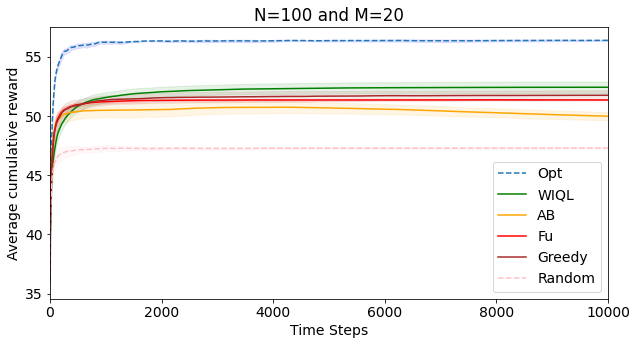}
  \caption{Average reward accumulated when only $20$ arms out of $100$ arms can be chosen at each iteration.}
  \label{fig:e2n100m20}
\end{subfigure}
\caption{Comparing various algorithms for Mentoring Instructions example. The results are averaged over $30$ trials.}
\label{fig:e2}
\end{figure*}

\paragraph{Results.} Figure~\ref{fig:e4} compares the performance of our proposed method with the benchmarks. Similar to the \textit{Circulant Dynamics} example, we observe that WIQL performs better in terms of the average total reward, compared to the other benchmark algorithms for the \textit{Example with Restart} problem where the number of states is higher.

\subsubsection{Mentoring Instructions}
This RMAB problem is considered by~\cite{fu2019towards}. Here, each arm, representing a student is assumed to be in one of the ten states $\mathcal{S} = \{0,\ldots, 9\}$ as any point of time. At each time step, only $M$ mentors are available and thus, action $a=1$ can be taken only on $M$ arms, which would possibly improve the state of the student. The higher index of the state implies better reward (state $9$ represents the best study level and state $1$ represents the worst study level). In particular, the reward at a state $Z\in\mathcal{S}$ is assumed to be $R^Z=\sqrt{\frac{Z}{10}}$. The transition probabilities for each action $a\in\{0,1\}$ are represented as a $|\mathcal{S}|\times |\mathcal{S}|$ matrix, where $p^1= 0.7$, $q^1=0.3$, $p^0= 0.7$, and $q^0=0.3$. 
\begin{equation*}
\mathcal{P}^a = 
\begin{pmatrix}
q^a & p^a &  &  &  \\
q^a & 0 & \ddots &  &  \\
  & q^a & \ddots & p^a  &  \\
  &  & \ddots & 0 & p^a \\
  &  & & q^a & p^a 
\end{pmatrix}
\end{equation*}

\paragraph{Results.} Figure~\ref{fig:e2} compares the performance of our proposed method with the benchmarks. We observe that WIQL performs marginally better in terms of the average total reward, compared to the other benchmark algorithms for the \textit{Mentoring Instructions} problem. However, unlike the previous example, the margin of improvement in performance is very low. A possible reason is the number of states being much more than that of the earlier example, which takes more time for the algorithms to learn the best policy. However, this example clearly shows that Random policy may perform even worse when the number of states is as high as 10 states.
\end{document}